# Towards Efficient Motion Planning for UAVs: Lazy A* Search with Motion Primitives

Wentao Wang
Department of Computer Science
University of Southern California
LA, USA
wwang047@usc.edu

Yi Shen
Robotics Department
University of Michigan
MI, USA
shenrsc@umich.edu

Kaiyang Chen
Department of Electrical and Computer Engineering
University of Illinois Urbana-Champaign
Champaign, USA
kc68@illinois.edu

Kaifan Lu
Department of Computer Science
University of Southern California
LA, USA
kaifanlu@usc.edu

*Abstract*—Search-based motion planning algorithms have been widely utilized for unmanned aerial vehicles (UAVs). However, deploying these algorithms on real UAVs faces challenges due to limited onboard computational resources. The algorithms struggle to find solutions in high-dimensional search spaces and require considerable time to ensure that the trajectories are dynamically feasible. This paper incorporates the lazy search concept into search-based planning algorithms to address the critical issue of real-time planning for collision-free and dynamically feasible trajectories on UAVs. We demonstrate that the lazy search motion planning algorithm can efficiently find optimal trajectories and significantly improve computational efficiency.

## I. Introduction

The demand for UAVs has surged across various domains, including search and rescue operations, environmental monitoring, and precision agriculture. Central to the success of these applications is the ability to plan and execute safe, efficient, and dynamically feasible trajectories in real-time. To achieve this, various motion planning algorithms have been developed, including search-based algorithms [1] [2], sampling-based algorithms [3], optimization-based algorithms [4] [5], and learning-based algorithms [6] [7].

Search-based motion planning algorithms have shown potential in reliable trajectory planning. The search process typically involves two main actions: node exploration and edge evaluation. This means adding a new potential node to the current search queue and evaluating the edge between the new node and its parent node. However, one of the main issues with conventional search-based algorithms for UAVs [1] [2] is the extensive time to perform edge evaluation.

To address these limitations, we proposed a novel approach that integrates lazy search with motion primitives A* algorithm. The lazy search technique delays the full edge evaluation until it is necessary, which significantly reduces computational overhead. By leveraging motion primitives, the algorithm can quickly generate dynamically feasible trajectories without exhaustively searching the entire high-dimension space.

The remainder of this paper is organized as follows: Section 2 reviews related work in motion planning. Section 3 outlines the motion planning problem from a search perspective. Section 4 describes the proposed Lazy A* Search Algorithm with Motion Primitives in detail. Section 5 presents the experimental setup and results. Section 6 concludes the paper and outlines directions for future research.

## II. Related Works

### A. The A* Algorithm and Its Variants

The A* algorithm [8] is a widely used search algorithm in path finding and graph traversal, known for its optimality and completeness. It uses a heuristic function that combines two costs to guide the search: the actual cost to reach a node and the estimated cost from the node to the goal. The Theta* algorithm [9] is an any-angle path planning algorithm based on A* algorithm. It allows for direct line-of-sight connections between nodes to reduce path length and improve efficiency. The D* algorithm [10] is designed for dynamic environment. It deals with dynamic obstacles by real time changing its edge's weights to efficiently replan path in response to changes in the environment.

While these algorithms are effective in their designed scenarios, they do not consider the dynamics of the robot, requiring post-processing trajectories to become traversable. To address this limitation, motion primitive A* algorithms [2] were developed. These approaches integrate motion primitives, which are segments of feasible trajectories into the A* search. By enforcing dynamic constraints on motion primitives, these algorithms generate trajectories that are immediately feasible for robot to traverse and eliminate the need for post-processing.

## B. Lazy Search Algorithms

Lazy search algorithms have explored various techniques to enhance efficiency in large and complex search spaces by deferring evaluations until necessary. The Lazy A* algorithm [11] delays edge evaluations until necessary. The Lazy Theta* [12] defers line-of-sight checks to improve pathfinding efficiency. Another significant contribution is Lazy Shortest Path [13] which postpones the evaluation of edge costs until they are crucial for determining the shortest path. These approaches have shown notable improvements in path planning problems by reducing unnecessary computations and enabling faster, scalable search solutions.

## III. PROBLEM FORMULATION

Let $x(t) \in X \subset \mathbb{R}^{3n}$ be the state of the UAV, which includes the position and its $n-1$ derivatives $(x, y, z, \dot{x}, \dot{y}, \dot{z}, \ddot{x}, \ddot{y}, \ddot{z}, \dots)$ in three-dimension space. Define $X_{free} \subset X$ as the free region in the state space. The free region $X_{free}$ indicates not only the obstacle-free positions $P_{free}$ but also constraints on the system's dynamics such as maximum velocity $v_{max}$, maximum acceleration $a_{max}$, and higher order derivatives for each axis. Thus $X_{free} \coloneqq P_{free} \times [-v_{max}, v_{max}]^3 \times [-a_{max}, a_{max}]^3 \times \dots$. Define the obstacle region as $X_{obs} \coloneqq X \setminus X_{free}$.

As outlined in [5], the differential flatness of quadrotor systems allows us to transform the complex, nonlinear dynamics into a set of simpler equations using a set of flat outputs. The flat outputs are chosen as the position coordinates in three-dimensional space $(x, y, z)$ and the yaw angle $\psi$. These flat outputs are sufficient to describe the full state and control inputs of the quadrotor. For many applications, the yaw angle $\psi$ and the corresponding dynamics it describes are not critical. Thus, for quadrotors, we generally focus on flat outputs $(x, y, z)$, which is sufficient to capture the key dynamics, including roll, pitch, velocity, acceleration, and jerk.

Define the control input as $u(t) \in U \coloneqq [-u_{max}, u_{max}] \subset \mathbb{R}^3$. Given the state $x(t)$, the dynamic model of a UAV can be written as a linear system as

$$\dot{x} = Ax + Bu$$
$$A = \begin{bmatrix} 0 & I_3 & 0 & \cdots & 0 \\ 0 & 0 & I_3 & \cdots & 0 \\ \vdots & \ddots & \ddots & \ddots & \vdots \\ 0 & \cdots & \cdots & 0 & I_3 \\ 0 & \cdots & \cdots & 0 & 0 \end{bmatrix}, B = \begin{bmatrix} 0 \\ 0 \\ \vdots \\ 0 \\ I_3 \end{bmatrix} \quad (1)$$

Given the dynamic model, the current state variable $x(t)$, and the control input $u(t)$, the next state $x(t + \Delta t)$ over a small-time interval $\Delta t$ can be computed using numerical integration.

$$\dot{x}(t) = Ax(t) + Bu(t)$$
$$x(t + \Delta t) = x(t) + \dot{x}(t) \times \Delta t \quad (2)$$

Motion primitives are constructed to discretize the reachable state space for the current state. Following (2), it is evident that given a current state, control inputs can be sampled to determine all reachable state spaces. Instead of using the continuous control set $U$, a discretized control set $U_M \coloneqq \{u_1, u_2, \dots, u_M\} \subset U$ is sampled, where each control input $u \in \mathbb{R}^3$ defines a motion of short duration for the UAV. Thus, all reachable state space after a short time interval $\tau$ can be identified. The edge $e_{(i,j)}$ is defined as the trajectory from state $x_i$ to state $x_j$. And the trajectory $P$ is defined as a sequence of motion primitives. Then a cost function $C(x, e)$ based on a state and its edge can be defined to evaluate the cost of trajectory $J$.

$$J \coloneqq \{e_{(0,1)}, e_{(1,2)}, \dots, e_{(k-1,k)}\} \quad (3)$$

$$\text{The Cost of Trajectory } J = \sum_{i=0}^{k} C(x_i, e_{i,i+1}) \quad (4)$$

We now define the problem of motion planning as follows: Given an initial state $x_0 \in X_{free}$ and a goal region $X_{goal} \in X_{free}$, find a trajectory $J = \{e_{(0,1)}, e_{(1,2)}, \dots, e_{(k-1,k)}\}$ such that:

$$\min \sum_{t=0}^{T} C(x(t), e_{t,t+1})$$
$$s.t. \quad \dot{x}(t) = Ax(t) + Bu(t) \quad \forall t \in [0, T] \quad (5)$$
$$x(0) = x_0 \quad x(T) \in X_{goal}$$
$$x(t) \in X_{free} \quad u(t) \in U \quad \forall t \in [0, T]$$

In the remainder, we denote the optimal cost form an initial state $x_0$ to a goal region $X_{goal}$ as $C^*$. Given the three-dimensional control of the control space $U$, search-based algorithms, which discretize $U$ using motion primitives, are efficient and resolution complete.

## IV. LAZY A* SEARCH WITH MOTION PRIMITIVES

In this section, we discuss an approach to improve motion primitive sampling efficiency. Then, we will introduce our Lazy Algorithm, which incorporates the enhanced motion primitives.

### A. Motion Primitives

The construction of motion primitives aims to discretize the continuous state space into discrete states. These primitives are derived by sampling the control set $U$ and applying each constant control sample to the current state $x(t)$ for a duration $\tau$, then calculating the next state $x(t + \tau)$ according to (2).

One method to generate the control sample set $U_m$ involves uniformly sampling along each control axis $[-u_{max}, u_{max}]$. However, uniformly sampling along each control dimension does not ensure uniform motion primitive samples in the state space. To address this issue, we propose sampling control along each control axis $[-u_{max}, u_{max}]$ using a normal distribution $u_m \sim N(0, 1)$. This method achieves a more uniform distribution of state space samples.

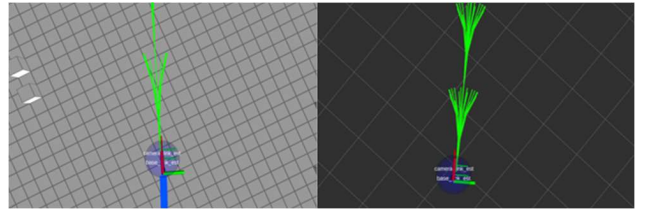

Fig. 1. Comparison of control sampling methods - Left: uniform control sampling; Right: control sampling in normal distribution.

To assess the efficacy of different control sampling strategies, we introduced two evaluation metrics:

- The ratio of useful samples to total samples $\alpha$: Control samples are considered one useful sample if the motion primitive endpoints lie within a distance of 0.1 meters. The ratio $\alpha$ is calculated by dividing the number of useful samples by the total number of samples in the control set $U_m$.

- The distance from each motion primitive endpoint to its nearest neighbor $L$ in centimeters: It evaluates the distribution of motion primitive endpoints. A larger $L$ indicates that the motion primitives cover a greater state space, thereby demonstrating stronger ability to explore.

TABLE I.  COMPARISION OF SAMPLING METHODS

|   | Random Samples | Uniform Sampling | Samples in Normal Distribution |
|---|---|---|---|
| $\alpha$ | 60% | 84% | **94%** |
| $L$ | 0.049 | 0.069 | **0.079** |

The results demonstrate that non-uniform control samples, those drawn from a normal distribution, significantly enhance sampling efficiency and exploration capabilities.

*B. The Algorithm*

The algorithm is specifically designed to delay edge (motion primitive) evaluation, thereby minimizing unnecessary computation effort. The cost function $C(x, e)$ described in (5) comprises four components:

- The $g$ cost: the actual distance from the start node to the current node.
- The $h$ cost: the heuristic estimate of the distance from the current node to the goal node.
- The control cost: the absolute value of the control input, aimed at reducing control effort.
- The obstacle cost: it assesses the proximity to obstacles. The objective is to ensure that the planned trajectory maintains a safe distance from obstacles.

Evaluating the obstacle cost is computationally intensive, as it requires querying the Euclidean Signed Distance Field (ESDF) map [14] to determine the distance the nearest obstacle. For an edge $e$, it is essential not only to evaluate the obstacle cost at the start and end points but also to assess the obstacle cost at intermediate points along the edge to ensure collision-free edges.

As shown in Fig. 1, we define a "fully evaluated edge" as one for which the $g$, $h$, control, and the obstacle costs are assessed at the start node, end node and all nodes along the edge. Conversely, a "partially evaluated edge" is one where these costs are evaluated only at the start and end nodes.

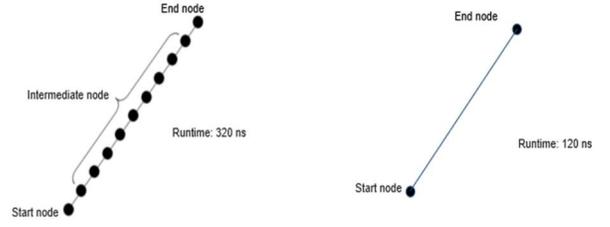

Fig. 2. Comparison of Fully Evaluated Edge (left) and Partially Evaluated Edge (right) – The fully evaluated edge checks all intermediate nodes with a runtime of 320 ns, whereas the partially evaluated edge assessed them only at the start and end nodes with a runtime of 120 ns.

We now present the pseudocode for our Lazy A* Search Algorithm with Motion Primitives. The code structure is based on the classic A* algorithm, with added requirement of verifying the edge state evaluation state each time a new node is popped out for exploration.

---

**Algorithm 1** Lazy A* Search with Motion Primitives

**Function** $get\_parent(node\ s)$
1:    Find the parent node $p$ of node $s$
2:    **return** node $p$

**Function** $get\_edge(node\ s_1, node\ s_2)$
1:    Find the edge $e$ from node $s_1$ to $s_2$
2:    **return** edge $e$

**Function** $is\_fully\_evaluated(edge\ e)$
1:    **if** $e$ is fully evaluated **then**
2:       **return** true
3:    **end if**
4:    **return** false

**Function** $reconstruct\_trajectory(node\ s)$
1:    Trajectory $J = \emptyset$
2:    **while** $parent(s) \neq s$
3:       Prepend node $s$ to trajectory $J$
4:       $s = parent(s)$
5:    **end while**
6:    **return** trajectory $J$

**Function** $find\_neighbors(node\ s)$
1:    $S = \emptyset$
2:    **for** each control sample $u \in U_m$ **do**
3:       apply control $u$ to node s and compute the resulting node $s'$ after time $\tau$
4:       $parent(s') = s$
5:       $S = S \cup \{s'\}$
6:    **end for**
7:    **return** S

**Function** $is\_collision\_free(node\ s)$
1:    **if** node $s$ is collision-free **then**
2:       **return** true
3:    **end if**
4:    **return** false

**Function** $is\_collision\_free(edge\ e)$
1:    **if** intermediate and end nodes on $e$ is collision-free **then**
2:       **return** true
3:    **end if**
4:    **return** false

**Function** $partially\_evaluate(edge\ e)$

1: Compute the $g$, $h$, control, and the obstacle costs at the start and end node of edge $e$
2: **return** the total sum of all costs

**Function** $fully\_evaluate(edge\ e)$

1: Compute the $g$, $h$, control, and the obstacle costs at the start node, end node and all nodes along edge $e$
2: **return** the total sum of all costs

---

**Algorithm 1** Lazy A* Search with Motion Primitives

**Input**: Initial state $x_0$, goal region $X_{goal}$, map $\phi$
**Output**: Trajectory $J$ from $x_0$ to $X_{goal}$ (if found)

1: OPEN LIST $= \emptyset$, VISITED LIST $= \emptyset$
2: Insert node $s_0 = (0, x_0)$ into OPEN LIST
3: **while** OPEN LIST $\neq \emptyset$ **do**
4:   Pop out the lowest-cost node $s = (c, x)$ from OPEN LIST
5:   $p = get\_parent(s)$, $e = get\_edge(p, s)$
6:   **if** $is\_fully\_evaluated(e)$ **then**
7:     **if** $s \in X_{goal}$ **then**
8:       **return** $reconstruct\_trajectory(s)$
9:     **end if**
10:     Let $S = find\_neighbors(s)$
11:     **for** each $s' \in S$ **do**
12:       **if** $s' \in$ VISITED LIST **then**
13:         **continue**
14:       **end if**
15:       **if** $is\_collision\_free(s')$ **then**
16:         Let $e' = edge(s, s')$
17:         $c' = partially\_evaluate(e')$
18:         Insert $s' = (c', x')$ into OPEN LIST
19:         Insert $s' = (c', x')$ into VISITED LIST
20:       **end if**
21:     **end for**
22:   **else**
23:     $c'' = fully\_evaluate(e)$
24:     Update the cost of node $s$ to $c''$
25:     **if** $is\_collision\_free(e)$ **then**
26:       Insert $s = (c'', x)$ back into OPEN LIST
27:     **end if**
28:   **end if**
29: **end while**

## C. Algorithm Analysis

### 1) The resulting search trees

The search tree of Motion Primitive A* Search involves thorough evaluation of each edge (motion primitive) during node expansion, resulting in a fully explored search tree with precise cost values. In contrast, Lazy A* Search with Motion Primitives defers the full evaluation of edges until necessary and only evaluates them when they are about to expanded. This leads to a smaller and less explored search tree, making it well-suited for large search spaces where many potential paths exist.

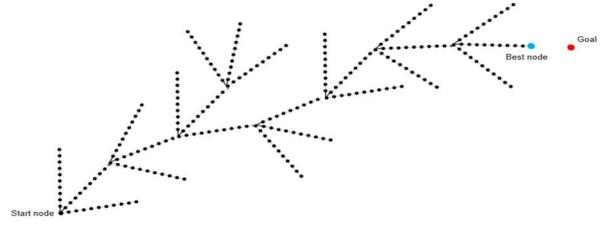

Fig. 3. Search Tree of Motion Primitive A* Algorithm

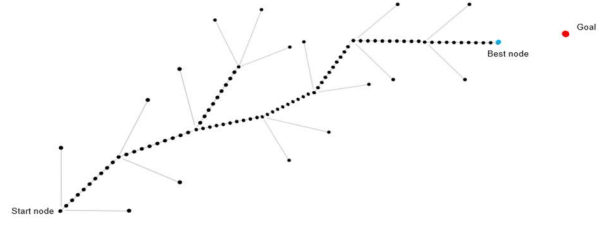

Fig. 4. Search Tree of Lazy A* Search with Motion Primitives

### 2) Proof of trajectory optimality

We assert that when a node is fully evaluated, its cost is greater than or equal to its partially evaluated cost. Assume the algorithm has found a path to the goal node $G$ with a cost $C^*$. By the time node $G$ is fully evaluated, all nodes, regardless of their evaluation status, with cost lower than $C^*$ have already been expanded. If a cheaper path to $G$ existed, the algorithm would have expanded node $G$ with a lower cost before finding the current path with cost $C^*$. Thus, we claim that the first path returned by the algorithm is the optimal trajectory.

## V. EXPERIMENTS

We conducted a series of experiments to evaluate the performance of the UAV both in simulation and real-world scenarios. The real-world experiment is conducted in a forest environment as illustrated in Fig. 5.

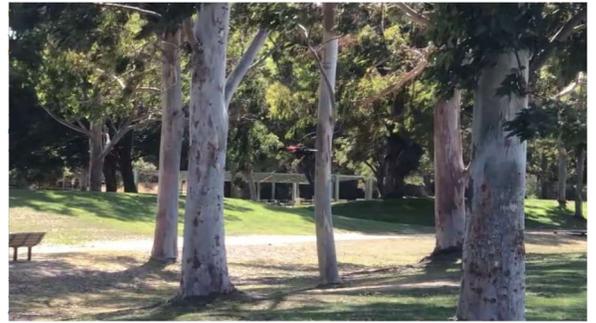

Fig. 5. Real-world Experiment Environment

To comprehensively assess the effectiveness of our planning algorithm, we established three key performance metrics.

- Average planning time $T$ in millisecond: This metric measures the time taken by the algorithm to generate a feasible trajectory from the initial position to the goal.

- Average number of node expansions $N$: This metric quantifies the number of nodes the algorithm expands during the planning process. It reflects the computation efficiency and the ability to explore the search space.

- Average distance from the best node to goal node $D$ in meters: In many instances, the trajectory generated by the planning algorithm does not exactly reach the designated end node. This metric measures the distance from the closest node in the generated path to the goal region. It serves as a metric to assess trajectory quality.

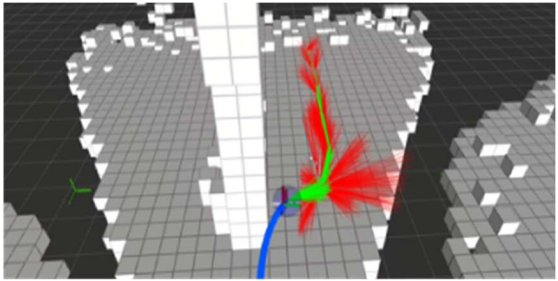

Fig. 6. Lazy A* Search with Motion Primitives – Red edges are partially evaluated edges and green edges are fully evaluted edges

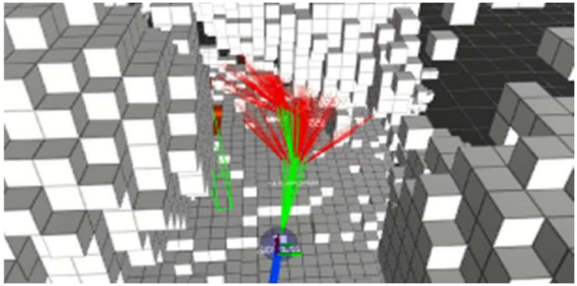

Fig. 7. Lazy A* Search with Motion Primitives in Complex Environment

We compare the performance metrics of three trajectory planners: TGK-Planner [3], Motion Primitive A* (MP-A*) [2], and the Lazy A* Search with Motion Primitives (Lazy).

The TGK planner utilizes a sampling-based path planner combined with trajectory optimization to produce smooth and feasible trajectories for UAVs, serving as a strong baseline.

The comparison of TGK, MP-A*, and Lazy algorithms shows that the Lazy algorithm is the most efficient. Although it explores a greater number of nodes, it achieves this with lower planning time. Further, the trajectory quality of the Lazy algorithm is comparable to that of MP-A* and TGK.

TABLE II.   COMPARISION OF MP-A* AND LAZY ALGORTHMS

|  | $T$ | $N$ | $D$ |
|---|---|---|---|
| **TGK** | 8.78 | - | 2.27 |
| **MP-A*** | 6.82 | **973** | 2.31 |
| **Lazy** | **3.87** | 1254 | **2.25** |

## VI. Conclusion

In this paper, we introduce a novel online motion planning algorithm, the Lazy A* Search with Motion Primitives. We validate our algorithm in both simulated and real-world challenging tasks. Experiments demonstrate that our algorithm significantly reduces planning time, while achieve comparable trajectory quality to that of MP-A* and TGK. In the future, we plan to explore machine learning method for edge evaluation and challenge our algorithm on large-scale problems.